\documentclass[a4paper, 10 pt, conference]{ieeeconf}

\IEEEoverridecommandlockouts                              % This command is only needed if 
\usepackage{xcolor}
\usepackage{graphicx}
\usepackage{xspace}
\usepackage{listings}
\usepackage{siunitx}
\usepackage{booktabs}

\lstdefinelanguage{pseudocode}{
    basicstyle=\notsotiny,
	keywordstyle=\bf \notsotiny,
	mathescape=true,
	tabsize=20,
	xleftmargin=4.0ex,
	basicstyle=\normalsize,
	sensitive=false,
	columns=fullflexible,
	keepspaces=false,
	basewidth=0.05em,
	moredelim=[il][\rm]{//},
	moredelim=[is][\sf \figuresize]{!}{!},
	moredelim=[is][\bf \figuresize]{*}{*},
	keywords={automaton, algorithm, and, 
		break,
		choose,const,continue, components,
		discrete, do,
		eff, external,else, elseif, evolve, end, each, exit,
		fi,for, forward, from, find, 
		hidden,
		in,input,internal,if,invariant, initially, imports,
		let,
		mode,
		or, output, operators, od, of,
		pre,
		return,
		such,satisfies, stop, signature, simulation, sample,
		trajectories,trajdef, transitions, that,then, type, types, to, tasks,
		variables, vocabulary, 
		when,where, with,while},
	emph={set, seq, tuple, map, array, enumeration},   
	literate=
	{(}{{$($}}1
	{)}{{$)$}}1
	{\\in}{{$\in\ $}}1
	{\\preceq}{{$\preceq\ $}}1
	{\\subset}{{$\subset\ $}}1
	{\\subseteq}{{$\subseteq\ $}}1
	{\\supset}{{$\supset\ $}}1
	{\\supseteq}{{$\supseteq\ $}}1
	{\\forall}{{$\forall$}}1
	{\\le}{{$\le\ $}}1
	{\\ge}{{$\ge\ $}}1
	{\\gets}{{$\gets\ $}}1
	{\\cup}{{$\cup\ $}}1
	{\\cap}{{$\cap\ $}}1
	{\\langle}{{$\langle$}}1
	{\\rangle}{{$\rangle$}}1
	{\\exists}{{$\exists\ $}}1
	{\\bot}{{$\bot$}}1
	{\\rip}{{$\rip$}}1
	{\\emptyset}{{$\emptyset$}}1
	{\\notin}{{$\notin\ $}}1
	{\\not\\exists}{{$\not\exists\ $}}1
	{\\ne}{{$\ne\ $}}1
	{\\to}{{$\to\ $}}1
	{\\implies}{{$\implies\ $}}1
	{<}{{$<\ $}}1
	{>}{{$>\ $}}1
	{=}{{$=\ $}}1
	{~}{{$\neg\ $}}1
	{|}{{$\mid$}}1
	{'}{{$^\prime$}}1
	{\\A}{{$\forall\ $}}1
	{\\E}{{$\exists\ $}}1
	{\\/}{{$\vee\,$}}1
	{\\vee}{{$\vee\,$}}1
	{/\\}{{$\wedge\,$}}1
	{\\wedge}{{$\wedge\,$}}1
	{=>}{{$\Rightarrow\ $}}1
	{->}{{$\rightarrow\ $}}1
	{<=}{{$\Leftarrow\ $}}1
	{<-}{{$\leftarrow\ $}}1
	{~=}{{$\neq\ $}}1
	{\\U}{{$\cup\ $}}1
	{\\I}{{$\cap\ $}}1
	{|-}{{$\vdash\ $}}1
	{-|}{{$\dashv\ $}}1
	{<<}{{$\ll\ $}}2
	{>>}{{$\gg\ $}}2
	{||}{{$\|$}}1
	{[}{{$[$}}1
	{]}{{$\,]$}}1
	{[[}{{$\langle$}}1
	{]]]}{{$]\rangle$}}1
	{]]}{{$\rangle$}}1
	{<=>}{{$\Leftrightarrow\ $}}2
	{<->}{{$\leftrightarrow\ $}}2
	{(+)}{{$\oplus\ $}}1
	{(-)}{{$\ominus\ $}}1
	{_i}{{$_{i}$}}1
	{_j}{{$_{j}$}}1
	{_{i,j}}{{$_{i,j}$}}3
	{_{j,i}}{{$_{j,i}$}}3
	{_0}{{$_0$}}1
	{_1}{{$_1$}}1
	{_2}{{$_2$}}1
	{_n}{{$_n$}}1
	{_p}{{$_p$}}1
	{_k}{{$_n$}}1
	{-}{{$\ms{-}$}}1
	{@}{{}}0
	{\\delta}{{$\delta$}}1
	{\\R}{{$\R$}}1
	{\\Rplus}{{$\Rplus$}}1
	{\\N}{{$\N$}}1
	{\\times}{{$\times\ $}}1
	{\\tau}{{$\tau$}}1
	{\\alpha}{{$\alpha$}}1
	{\\beta}{{$\beta$}}1
	{\\gamma}{{$\gamma$}}1
	{\\ell}{{$\ell\ $}}1
	{\\TT}{{\hspace{1.5em}}}3        
}

\lstdefinelanguage{pseudocodeNums}[]{pseudocode}
{
	numbers=left,
	numberstyle=\tiny,
	stepnumber=2,
	numbersep=4pt
}

\newcommand{\A}{\mathcal{A}}
\newcommand{\X}{\mathcal{X}}
\newcommand{\U}{\mathcal{U}}
\renewcommand{\P}{\mathcal{P}}
\newcommand{\brake}{\mathit{brake}}
\newcommand{\track}{\mathit{trackspeed}}

\newcommand{\reach}[1]{\mathsf{Reach}_{#1}}

\newcommand{\vx}{\bf{x}}

\usepackage{amsmath}
\usepackage{amsfonts}
\usepackage{todonotes}
\usepackage{tabularx}
\usepackage{caption}
\usepackage{subcaption}
\usepackage{algorithm}
\usepackage{algorithmic}
\usepackage{cite}
\usepackage{hyperref}

\title{\LARGE \bf
Online Monitoring for Safe Pedestrian-Vehicle Interactions
}

\newcommand{\decisionmodule}{\textit{DM}\xspace}
\newcommand{\piemodule}{\textit{PIE}\xspace}

\newcommand{\reachmodule}{\textit{OPRA}\xspace}

\let\svtodo\todo\renewcommand\todo[1]{\svtodo[inline]{#1}}

\author{Peter Du, Zhe Huang$^\dagger$, Tianqi Liu$^\dagger$, Tianchen Ji$^\dagger$, Ke Xu$^\dagger$, Qichao Gao$^\dagger$,\\ 
Hussein Sibai, Katherine Driggs-Campbell, and Sayan Mitra% <-this % stops a space
\thanks{*This research is supported by research grants from the National Science Foundation (FMITF:1918531) and the United States Air Force Office of Scientific Research (FA9550-17-1-0236). We are grateful to the Illinois Center for Autonomy, Coordinated Science Lab, and AutonomouStuff for supporting our experimental infrastructure.}% <-this % stops a space
\thanks{$^\dagger$ Z. Huang, T. Liu, T. Ji, K. Xu, and Q. Gao contributed equally.}% <-this % stops a space
\thanks{P. Du, Z. Huang, T. Liu, T. Ji, Q. Gao, H. Sibai, K. Driggs-Campbell, and S. Mitra are with the Department of Electrical and Computer Engineering at the University of Illinois at Urbana-Champaign. email: \{peterdu2,zheh4,tliu51,tj12,qgao10,sibai2,krdc,mitras\}@illinois.edu}%
\thanks{K. Xu is with the Department of Computer Science at the University of Illinois at Urbana-Champaign. email: kexu6@illinois.edu}%
}

\renewcommand{\baselinestretch}{.95}

\begin{document}

\maketitle
\thispagestyle{empty}
\pagestyle{empty}

%%%%%%%%%%%%%%%%%%%%%%%%%%%%%%%%%%%%%%%%%%%%%%%%%%%%%%%%%%%%%%%%%%%%%%%%%%%%%%%%

\begin{abstract}

As autonomous systems begin to operate amongst humans, methods for safe interaction must be investigated.
We consider an example of a small autonomous vehicle in a pedestrian zone that must safely maneuver around people in a free-form fashion.
We investigate two key questions: How can we effectively integrate pedestrian intent estimation into our autonomous stack? 
Can we develop an online monitoring framework to give rigorous assurances on the safety of such human-robot interactions?
We present a pedestrian intent estimation framework that can accurately predict future pedestrian trajectories given multiple possible goal locations.
We integrate this into a reachability-based online monitoring and decision making scheme that formally assesses the safety of these interactions with nearly real-time performance (approximately $0.1s$).
These techniques are both tested in simulation and integrated on a test vehicle with a complete in-house autonomous stack, demonstrating safe interaction in real-world experiments. 
\end{abstract}

\section{Introduction}
Autonomous systems are quickly entering human dominated fields in the form of drones and self-driving cars, and becoming tangible technologies that will impact the human experience.
As these systems begin to share space and operate among humans, safety becomes a primary concern~\cite{bajcsy2018scalable,fisac2018probabilistically}.
% Recent breakthroughs that bring autonomous vehicles closer to reality have led to growing research interest in designing safe learning and control frameworks~\cite{}.
Despite many successful demonstrations of autonomous driving, safe interaction between autonomous vehicles and other road users remains an open problem~\cite{thornton2018value,chen2017evaluation,driggs2017integrating}.

We consider two key factors currently impeding safe interactive autonomy.
First, designing interactive controllers that consider predictions about human movement remains a challenge due to the high variability in human behavior~\cite{driggs2018robust}.
Second, providing formal guarantees for autonomous systems is challenging due to complexities introduced by many intertwined modules in the autonomous stack as well as multi-agent interactions~\cite{govindarajan2017data}.

One test case where this is particularly obvious is an autonomous shuttle scenario, where an autonomous vehicle operates on small roads and sidewalks and may encounter pedestrians through free-form interaction.
Such automated systems have been developed for campus settings~\cite{rodriguez2017safety,van2017automated} and pedestrian zones~\cite{eden2017,kummerle2015autonomous}.
These systems typically require and have heuristic decision-making for determining safe interaction (e.g., if an obstacle is within some preset distance, stop~\cite{nordhoff2018user}).
One common ``lesson learned'' is that interacting with pedestrians is difficult due to their unpredictability~\cite{van2017automated,trulls2011autonomous}.
As pedestrians are vulnerable agents and have very few constraints on their motion, determining methods for safe interaction is both critical for reducing risk of accidents and improving efficiency by limiting over-conservative performance.
Further, many studies state that an on-board safety driver or watchdog is necessary to monitor the system operation~\cite{van2017automated,kummerle2015autonomous}. 

In this work, we present an initial foray into integrating human motion prediction with an autonomous system's control framework as well as an online monitoring system.
Leveraging data-driven verification approaches to provide formal guarantees, we demonstrate how  DryVR, an offline reachability tool for hybrid dynamical systems~\cite{FanQMV:CAV2017}, can be used in an online setting to give a formal assessment of safety as the autonomous vehicle drives in close proximity to a human.
In short, we present the following contributions:
\begin{enumerate}
	\item \emph{Intent estimation.} We present a particle filter-based approach to predict human motion and provide future trajectories as well as estimate their likelihoods over multiple goals for each agent. 
	\item \emph{Safety monitoring.} We develop a predictive reachability analysis technique. This near-realtime, forward reachability analysis  allows the autonomous system to make risk-aware decisions, and can help designers assess safety of the overall system.
	\item \emph{System integration.} We demonstrate the feasibility of our efforts on a complete autonomous system which includes pedestrian tracking and intent prediction, localization, waypoint tracking control, and an online reachability analysis module. 
\end{enumerate}

The paper is organized as follows. In Section \ref{sec:background}, we present a literature review of works related to modeling pedestrian intent and online monitoring, as applied to autonomous systems.
Then, we present an overview of our experimental platform, in-house autonomous stack, and methodology for pedestrian intent estimation in Section \ref{sec:autonomy-modules}.
In Section \ref{sec:monitoring}, we discuss our approach for online monitoring via reachability.
We conclude with an analysis of our integrated system and discuss findings and future directions in Sections \ref{sec:results} and \ref{sec:discussion}.
%%%%%%%%%%%%%%%%%%%%%%%%%%%%%%%%%%%%%%%%%%%%%%%%%%%%%%%%%%%%%%%%%%%%%%%%%%%%%%%%
\section{Background}
\label{sec:background}
We present a brief literature review of techniques for modeling pedestrian interactions for use in decision-making and control frameworks as well as verification methods to provide guarantees for autonomous systems.

\subsection{Pedestrian Modeling}
% The development of fully autonomous vehicles will inevitably encounter human-robot interaction cases, especially the interaction between pedestrians and vehicles.
% Effective pedestrian trajectory prediction would benefit decision-making and control modules to keep driving efficiency with guarantee on safety. 
There are three main techniques used to predict pedestrian trajectories: data-driven approaches like neural networks, decision-making approaches like Markov models, and filtering approaches like Kalman filters.

Neural networks, especially Long Short Term Memory Networks (LSTM)\cite{alahi2016social, gupta2018social, zhang2019sr}, have been successfully implemented to model social interactions between humans in crowded scenarios by integrating neighbor information into pedestrian trajectory prediction. Yi et al. encoded pedestrian behavior into sparse displacement volumes, which are used as input and output of the Behavior-CNN they proposed~\cite{yi2016pedestrian}. These methods are built on the assumption that the observation data provides perfect pedestrian positions, so they may not guarantee robust performance under noisy measurements. 

% Currently data-driven methods are becoming popular as it is capable of modeling interaction between multiple pedestrians, but most of them are built upon the assumption that observation data provides ground truth information. The performance of neural networks is not guaranteed under noisy measurement conditions, whereas the filtering techniques are a good fit to real world applications. In addition, the current focus of our study is on interaction between pedestrian and vehicle. Only a single pedestrian is considered in the experiment to simplify the problem. The integration of human-human interaction and human-vehicle interaction will leave to future study.

There has also been plenty of research where pedestrian behavior modeling is treated as a sequential decision making problem \cite{karasev2016intent, bai2015intention, bandyopadhyay2013intention, activityforcasting}. Jump-Markov process is proposed in \cite{karasev2016intent} to model pedestrian behavior, which assumes that the goal is slowly time-varying and that the pedestrians are rational. Partially Observable Markov Decision Process (POMDP) \cite{bai2015intention} and Mixed Observability Markov Decision Process (MOMDP) \cite{bandyopadhyay2013intention} are used to incorporate intents as hidden variables. In different problem settings, intents may be defined as goals \cite{bandyopadhyay2013intention}, sub-goals \cite{bai2015intention, ikeda2013modeling} or a policy in a Markov Decision Process \cite{karasev2016intent}. In our study, the intent always refers to the desired goal of the pedestrian.

Filtering is one of the most common methods for state estimation in real world applications. Filtering can be applied to not only pedestrian tracking \cite{wang2007wlan, meuter2008unscented, wang2010modified} but also pedestrian movement prediction \cite{schneider2013pedestrian, goalOrientedPedesterianDetection, particke2018improvements, particke2018entropy}. Moreover, particle filtering integrated with intent hypotheses model pedestrian behavior as a multi-hypotheses dynamical system, and thus capture the switching feature of complicated pedestrian dynamics \cite{particke2018improvements, particke2018entropy}. Thus, our work is built on \cite{particke2018improvements} to provide effective pedestrian intent estimation given noisy measurements.

\subsection{Reachability Analysis and Safety Verification}
%\todo{Add additional references mentioned in ICRA reviews (posted on slack channel)}
%
The {\em reachability\/} problem asks, given a system model $\A$, an initial set of states $\Theta$, and a particular target state $\vx^*$, whether any behavior of $\A$ starting from $\Theta$ can reach $\vx^*$?
Reachability analysis has played a fundamental role in both control theory and formal methods for cyber-physical systems~\cite{alur95algorithmic,Tabuada:2009:VCH,lygeros2003dynamical,alex02hscc}. A key application of reachability analysis that is relevant for this paper is in {\em safety verification\/}: If the target $\vx^*$  is  defined as an {\em unsafe state}, then answers to the reachability question can determine whether the system $\A$ can become unsafe starting from $\Theta$. 

It is well-known that the reachability problem is {\em undecidable\/} for general dynamical and hybrid system models~\cite{henzinger95algorithmic}.
Researchers have developed algorithms for relaxations of the problem by considering (a)  finite time horizon behaviors of $\A$, (b) allowing false-positives in the answers, and (c) allowing answers with probabilistic guarantees.  
A series of breakthroughs have led to the creation of software tools and libraries that can perform effective reachability analysis. For example, SpaceEX can perform reachability analysis of linear hybrid systems with dozens of continuous dimensions~\cite{Spaceex};  HyLAA has successfully verified sparse linear models with thousands of dimensions~\cite{bak2017hylaa,BakTJ19}. Nonlinear models with  hundreds of dimensions can now be verified with tools like C2E2~\cite{DMVemsoft2013,FanQM18}, Flow*~\cite{Flow}, and CORA~\cite{CORA16}.
All of these approaches are geared towards {\em off-line safety verification\/} and they have been applied to autonomous vehicles~\cite{FanQM18}, medical devices~\cite{ChenDS17}, and space operations~\cite{ChanM17CDC}.

Online reachability analysis can be used for predictive safety monitoring and decision making for an autonomous vehicle. In this setup, $\A$ is the model of the vehicle, and 
the initial set $\Theta$ is instantiated to be the current state estimate for this system (e.g., state of vehicle and pedestrians). Then  reachability analysis up to a {\em look-ahead\/} time horizon $T_{Look} >0$ can predict whether the vehicle is going to be unsafe within $T_{Look}$ time. For this analysis to be useful the reachable set computation has to finish before the next control decision, which is typically $10$-$100$ milliseconds. We will revisit these issues in Section~\ref{sec:monitoring}. 

The feasibility of online verification for  automatic lane change maneuvers on a Cadillac SRX was demonstrated in~\cite{Online-AlthoffD14,MathiasReachVehicle11}. This analysis used linearized approximations of the vehicle dynamics relative to the behaviors of all possible surrounding cars. In contrast, our reachability analysis and decision module uses pedestrian intents. Further, our data-driven reachability analysis approach can directly handle  nonlinear vehicle models.
%where as~\cite{Online-AlthoffD14}. 
%
%Finally, although the computing cycles available on the Cadillac SRX are not discussed in~\cite{Online-AlthoffD14}, we note that our on-board computers running  vision, positioning, and estimation algorithms severely constrained the network bandwidth and the cycles available for monitoring. 
%
%\hussein{talk about these citations}
In~\cite{VehiclePedesterianInteraction19}, a research platform that is similar to ours was presented, where uncertainty of the predicted behavior of the pedestrians is taken into account but the vehicle dynamics is not used for reachability.
%analysis to account for modeling and sensing errors.  
In a similar vein, \cite{ThreatAssesReach11} uses reachability analysis to check  safety of human-driven vehicles. 
%They use a driver and vehicle model to predict the behavior of the car under uncertainties. 
This work also uses linearized dynamics and does not  implement the solution on a full autonomous vehicle nor does it address pedestrian intent detection.
Several works use reachability for safe controller synthesis, typically assuming perfect sensor information within static  environments~\cite{BajcsyReachNavigation19,SafetyLearn17}. 
%Our approach differs in being focused on safety verification using reachability analysis while changing the mode of the car between $\mathit{brake}$ and $\mathit{cruise}$ accordingly to provide safety. 
%We use a fixed PID controller rather than synthesizing one. Last, we consider unknown and varying environment while estimating the behaviour of the pedestrians and accounting for sensing errors.
Finally, in \cite{InfuseReach18}, an approach similar to ours is followed in the context of safety relative to neighboring human driven vehicles. 
% %https://www.researchgate.net/publication/338754531_On_Infusing_Reachability-Based_Safety_Assurance_Within_Probabilistic_Planning_Frameworks_for_Human-Robot_Vehicle_Interactions}
%It adjusts the planned trajectory based on the safety result of the reachability analysis. It computes and caches reach sets offline and retrieves them online to check safety.  
%\cite{SafetyLearn17}, similar to \cite{BajcsyReachNavigation19}, focuses on safe online learning of controllers by checking the safety of different proposed control actions using reachability analysis with the Hamilton-Jacobi approach and applied it to drones.

%%%%%%%%%%%%%%%%%%%%%%%%%%%%%%%%%%%%%%%%%%%%%%%%%%%%%%%%%%%%%%%%%%%%%%%%%%%%%%%%
\section{Autonomy Modules}
\label{sec:autonomy-modules}

We describe our autonomous testbed in terms of the experimental hardware and the software components used in our autonomous stack.

\vspace{-0.3cm}
\begin{figure}[!h]
	\centering
	\includegraphics[width=0.98\columnwidth]{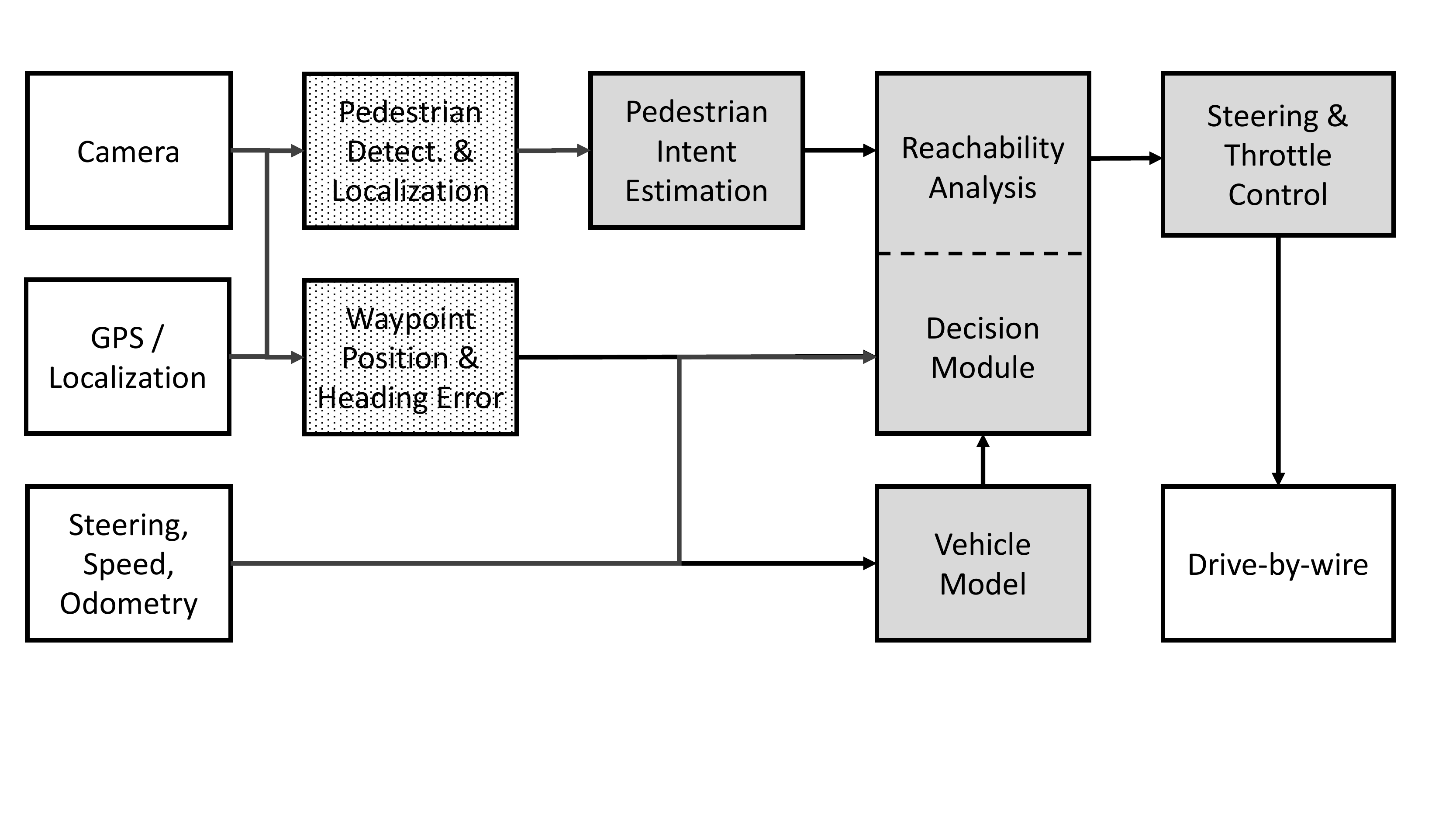}
	\captionsetup{justification=centering}
	\vspace{-0.8cm}
	\caption{System architecture. The dark gray modules constitute the key contributions of this work.}
% 	\vspace{-0.3cm}
% 	\vspace{-10pt}
		\label{fig:architecture}
\end{figure}
\vspace{-0.3cm}

\subsection{Vehicle Hardware and Experimental Setup}
\label{sec:vehicle-hardware}
\begin{figure}[!h]
	\centering
	\includegraphics[width=0.45\columnwidth]{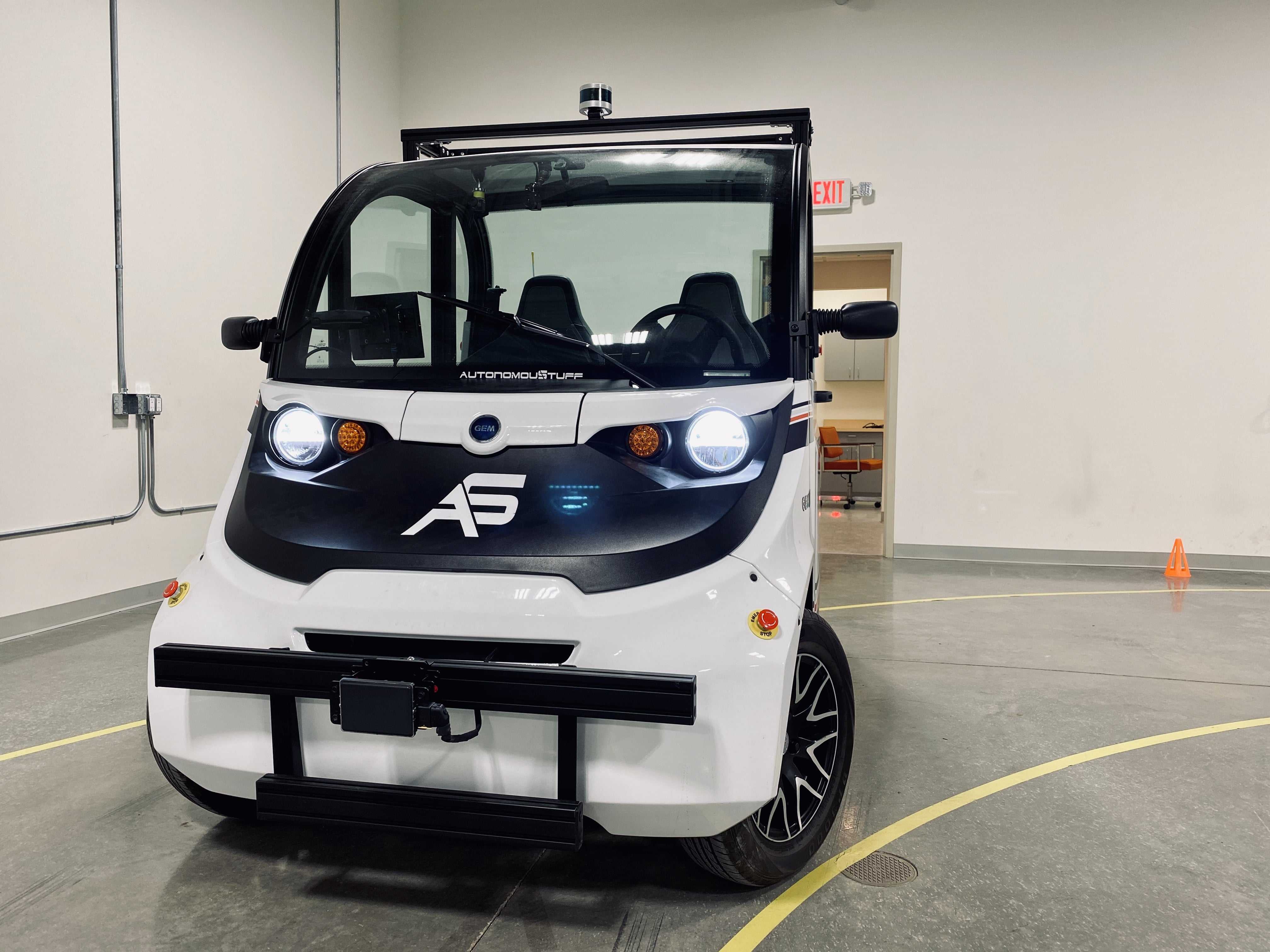}
	\includegraphics[width=0.45\columnwidth]{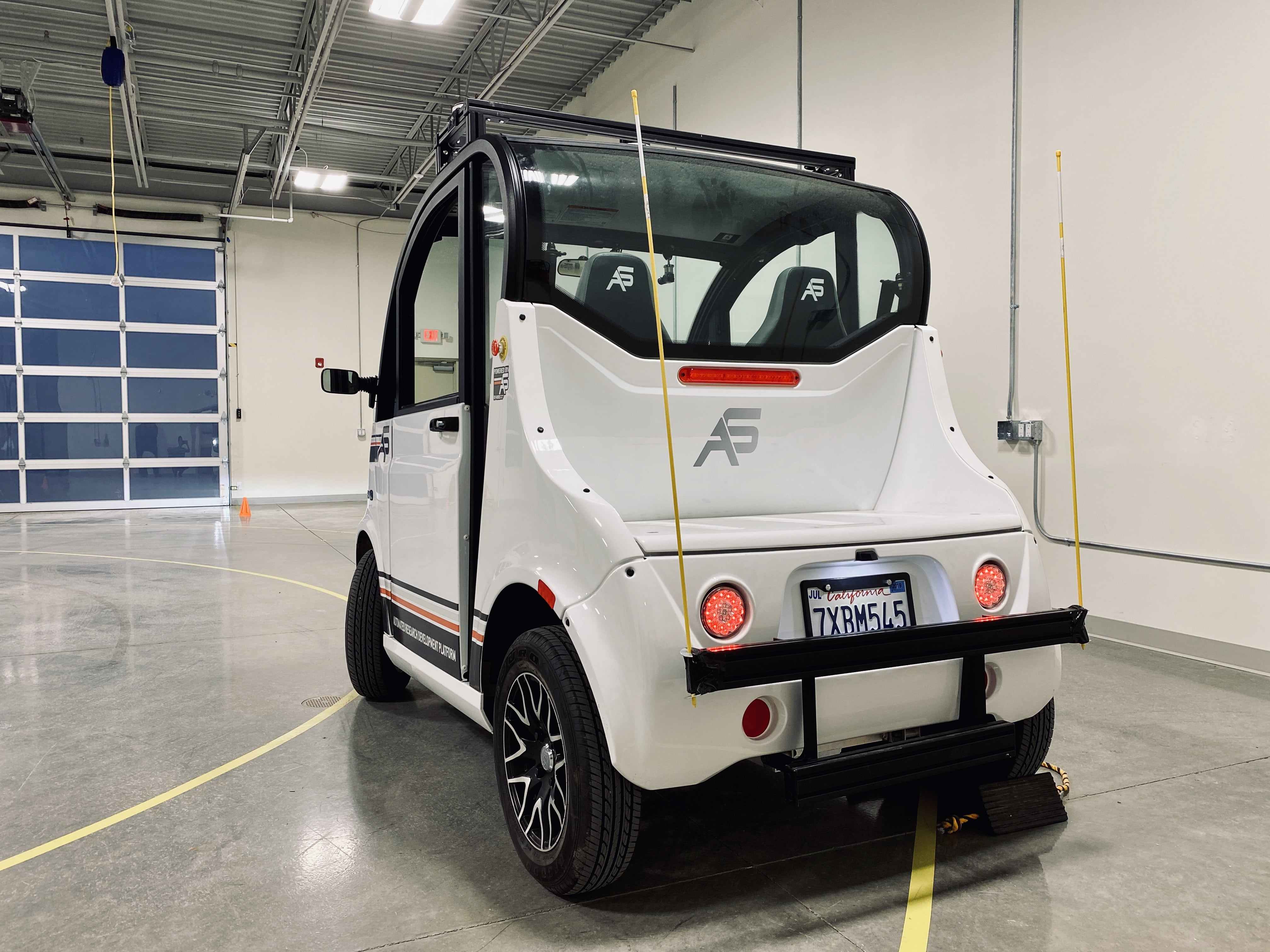}
	\caption{ Polaris GEM E2 in our testing arena.}
	\vspace{-0.3cm}
	\vspace{-10pt}
	\label{fig:polaris_gem}
\end{figure}

Our vehicle is a Polaris GEM electric car outfitted with sensing and computing hardware by AutonomouStuff (Figure~\ref{fig:polaris_gem}). The sensors include a Velodyne LIDAR, a Radar, GPS \& Inertial Measurement Units, and a single Mako G-319C forward facing camera. Our experimental arena is a 3,000 sq. ft. indoor open research space.

\subsection{Localization and Pedestrian Detection}
\label{pedestrian}

% With kalman filter part
We implemented a localization protocol that gives 2D global coordinates of the vehicle using Decawave ultra-wideband beacons and time-of-flight calculations. To improve the precision and reduce the errors caused by noise, Kalman filtering is applied on the position measurements received from Decawave. In our testing, the localization protocol can provide a 2D localization accuracy within 20 cm and message drop rates less than one per minute with position data update rates of 3Hz. Our indoor testing facility was unobstructed by walls and allowed for a clear line of sight among the Decawave chips.

\begin{figure}[!h]
	\centering
	\includegraphics[width=0.80\columnwidth]{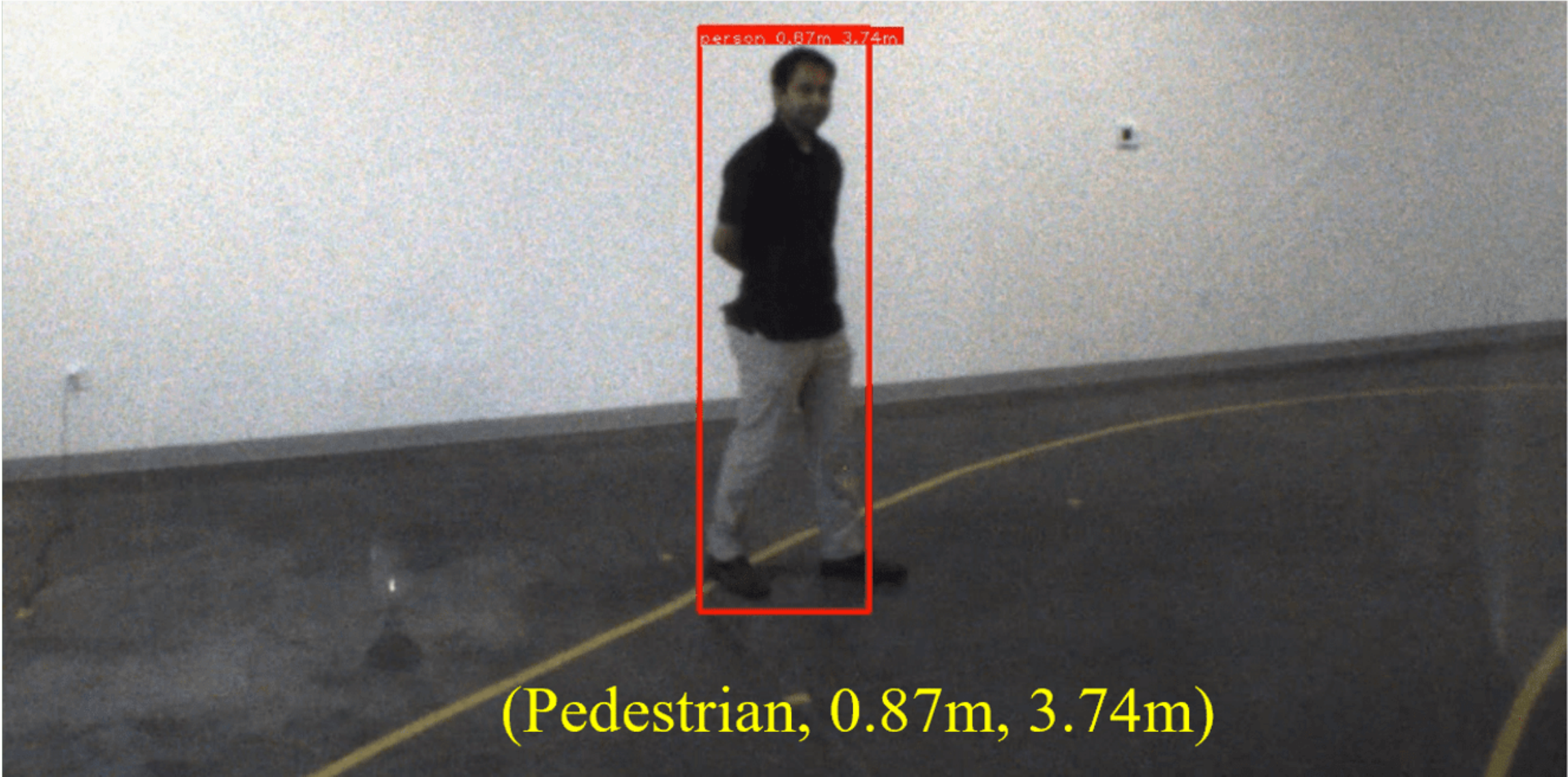}
	\captionsetup{justification=centering}
	\caption{Detected pedestrian and position estimates.}
% 	\vspace{-0.3cm}
% 	\vspace{-10pt}
		\label{fig:ped_detect}
\end{figure}

For pedestrian detection, we use YOLOv3: a state-of-the-art real-time object detection system. Once a pedestrian is detected on an input image, YOLOv3 places a bounding box around it. To estimate the longitudinal distance $y$ of the detected pedestrian with respect to the vehicle, we use the size of bounding box, focal length of the camera, and assumed width of the pedestrian ($0.47$ m). The same is done to estimate the lateral deviation $x$. The estimated $x$ and $y$-coordinates are converted from vehicle coordinates to global coordinates using position data from the vehicle's localization. Using this approach, we get an observed error on the pedestrian's location that does not exceed $0.5m$.

\subsection{Pedestrian Intent Estimation (\piemodule) Module}
\label{sec:pie}

The pedestrian intent module takes a pedestrian position measurement in the global coordinate frame and produces a future trajectory that ends at a predefined goal state. The predicted trajectory is then fed into the decision module ($\decisionmodule$) to assess the safety of the driving scenario and advise the controller accordingly.

The estimation algorithm assumes that the map information and all possible pedestrian intents are given as a set of possible goal locations. 
For example, in a crosswalk scenario, the pedestrian may have two possible intents. One is to cross the road, which is represented by an intent on the other side of the road. The other is to go alongside the road, for which the intent can be set as a location on the same side of the road as the pedestrian. Based on this assumption, the sequential measurements of the pedestrian are used to predict its intent.

An accurate pedestrian model is necessary to predict the intent. The Generalized Potential Field Approach (GPFA) is used to model the pedestrian behavior~\cite{particke2017pedestrian}.  
A set of sources are created to represent the obstacles in the map and to generate the repulsive force on the pedestrian, while a sink is set at goal locations to attract the pedestrian. The sum of forces is the acceleration, which is the control input $u_t$ to the state space model of the pedestrian, giving the following dynamical model: 
\begin{equation}\label{eq:ped_state_model}
    \begin{array}{rcl}
         x_{t} &=& F x_{t-1} + G u_{t-1} + w_{t}  \\
         y_{t} &=& H x_{t-1} +  v_{t} \\
    \end{array}
\end{equation}
where $F$ is the state transition matrix, $G$ is the control matrix, $x_{t}$ is the state vector (position and velocity) of the pedestrian at time $t$, $H$ is the measurement matrix, and $y_{t}$ is the measurement of the pedestrian's location.
It is assumed that the process noise $w_{t}$ and the measurement noise $v_{t}$ are Gaussian white noise.

% There can be multiple initial guesses of pedestrian intents in interactive scenarios between the pedestrian and the vehicle. 
As the transition model involves a highly nonlinear GPFA, we employ a Multi-hypothesis Particle Filter to estimate the likely intent~\cite{particke2018improvements}. 
Initially, each particle sampled will be assigned a hypothesis on pedestrian intent $I_i$. The number of intents $N$ considered in this paper is three, corresponding to either crossing the vehicle's path or heading parallel along the path in either direction. 
% represents the cases of going across and along the road), but the particle filter works the same for cases where $N_I > 2$. 
The probability of assigning different intents to a particle is the same, as $1/N_I$ for each intent. In the prediction step, each particle predicts the pedestrian position at the next time stamp based on the hypothesis of the intent to which it is assigned. In the update step, the weights of particles are updated based on the deviation of the position prediction from the measurement.

In contrast to the method presented in~\cite{particke2018improvements}, the weights of particles with the same intent hypothesis are summed to generate the probability distribution of different intents at that timestamp.
% The difference between the particle filter of this paper and \cite{particke2018improvements} is after the normalization of weights in the update step, the weights of particles with the same intent hypothesis are summed up to generate the probability distribution of different intents at that time stamp. 
Sequential importance re-sampling is implemented after the weights are updated. The intent with the highest probability is chosen to create the trajectory of the pedestrian in the future using GPFA based on the current measured position. 
In \cite{particke2018improvements}, the re-sampling step was not needed due to the robustness requirement for changes of the true intent of the pedestrian. 
However, no re-sampling would make the intent estimation vulnerable to sample degeneracy (i.e., only few particles with significant weights are kept throughout the intent estimation process), which leads to the loss of population nature of the particle filter. In fact, sequential importance re-sampling makes use of the quantity of particles to inherit the probability distribution inferred from the last round and causes no harm to the algorithm's robustness.
% The forward prediction of trajectory is the pedestrian part of the input to the online reachability analysis. 
Another difference is that though many of the model parameters are the same as in \cite{particke2017pedestrian, particke2018improvements}, no additional noise is manually added due to the sensor noise inherent in the measurement.
For online performance, we initially generate 200 particles for each intent (600 total).

\begin{figure}[t!]
    \hspace*{-0.5cm}
    \centering
    \includegraphics[width=0.90\columnwidth]{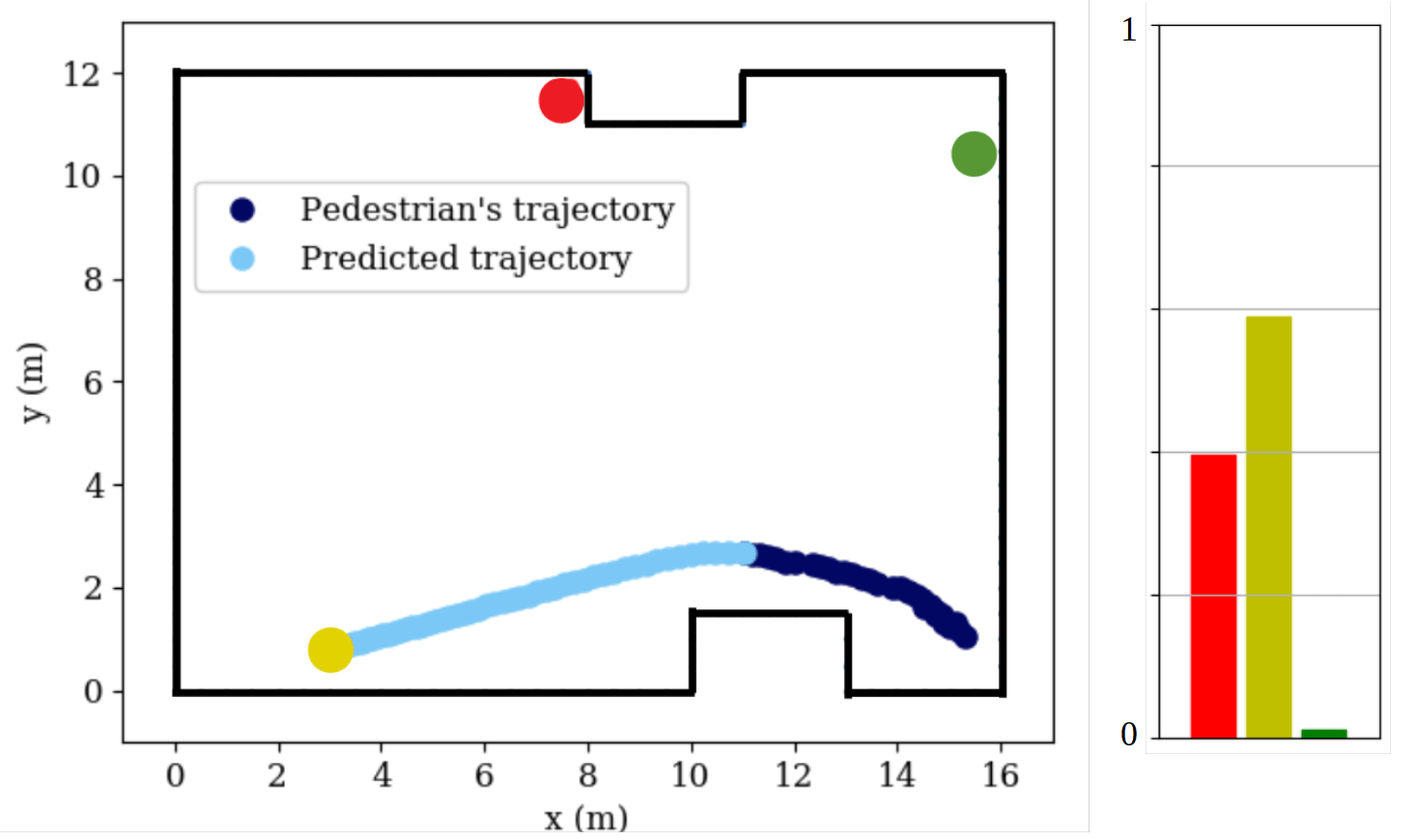}
    \captionsetup{justification=centering}
    \caption{Pedestrian intent estimation for scenario with three goals (red, yellow, green). \emph{Left:} Predicted trajectory. \emph{Right:} Probability distribution of pedestrian intents.}
    \label{fig:ped_intent}
    \vspace{-0.2cm}
\end{figure}

Ultimately, this module produces the estimated pedestrian trajectory as a sequence of location-time pairs $\tau = \{\langle p_1, t_1 \rangle,\ldots,\langle p_k, t_k \rangle\}$, where the final point is the estimated goal location. 
The algorithm runs at 5 Hz, which is sufficient for the online monitoring application. Figure \ref{fig:ped_intent} shows an example of the algorithm output on a public dataset \cite{majeckaedinburgh}. There are three possible intents in this case (shown in red, yellow, and green). The pedestrian trajectory is in dark blue and the predicted trajectory is in light blue. Our methodology is well tuned and validated on this public dataset \cite{majeckaedinburgh}, accurately predicting the intent and effectively producing the future trajectory.

\subsection{Velocity and Waypoint Following Control}
\label{sec:control}

% \begin{figure}[!h]
% 	\centering
% 	\includegraphics[width=0.85\columnwidth]{Figures/controller.png}
% 	\captionsetup{justification=centering}
% 	\vspace{-0.3cm}
% 	\caption{\scriptsize Steering and throttle controller.}
% 	\label{fig:controller}
% \end{figure}

We developed a control module to generate throttle and steering commands to navigate the vehicle autonomously. The control module is integrated into our ROS framework to interface with the other modules of the GEM vehicle platform as well as the PACMod drive-by-wire system (see Figure~\ref{fig:architecture}). We use two standard PID controllers, one for maintaining the vehicle at a desired speed $v_r$ (set by the user), and a second one for steering the vehicle through a sequence of predefined waypoints. 
% The inputs to the control module includes the reference signals $r(t)$ for speed and lateral deviation, as well as a safety indicator from the reachability-based decision module \decisionmodule. The outputs includes the throttle (acceleration/braking) $a$ and steering control $u$ inputs.
Spline curve fitting is used to get a planned path through the waypoints. The lateral deviation of the vehicle from the path is obtained at each timestep, and used to generate a steering velocity command with the PID steering velocity controller. 
%Vehicle speed control is done in a similar fashion. 
% The speed controller reads the current vehicle speed and acceleration commands are generated to track a reference speed. The speed controller takes an additional command from the reachability-based  decision module ($\decisionmodule$), which chooses the mode of operation between $\brake$ and $\track$ based on the the safety of the predicted trajectory of the car. When the current path is safe, the controller operates in the $\track$ mode the acceleration $a$ is the output of the PID speed controller. When the current path is unsafe, the controller operates in the $\brake$ mode and the acceleration $a$ is set to a constant $-5 m/s^2$. In both modes, the steering output $\phi$ is taken from the PID steering controller.

The speed controller reads the current vehicle speed and outputs acceleration commands to track the reference speed $v_r$. It takes an additional input from the reachability-based  decision module ($\decisionmodule$), which chooses the mode of operation between $\brake$ and $\track$, based on the the safety of the predicted behavior of the car. If the $\decisionmodule$ chose the $\track$ mode, the PID speed controller sets the acceleration $a$ of the vehicle to track $v_r$. If the $\decisionmodule$ chose the $\brake$ mode instead, the speed controller releases the throttle and applies the brakes to bring the vehicle to a stop. In both modes, the steering velocity output $u$ is taken from the PID steering controller to track the planned path.

In summary, the inputs to the control module include the state of the vehicle shown in Section~\ref{sec:monitoring}, reference constant speed $v_r$, lateral deviation from the planned path, as well as a chosen mode of operation from the reachability-based decision module \decisionmodule. The outputs include the throttle (acceleration/braking) $a$ and steering velocity control $u$.

\section{Online Predictive Reachability Analysis Module (\reachmodule)}
\label{sec:monitoring}
%In this paper, we consider a low-dimensional vehicle model $\A$ to predict the behavior of the real car for few seconds ahead of its current state. The model $\A$ is a tuple $\langle X, U, P, f, g \rangle$, where $X \subseteq \mathbb{R}^n$ is the set of possible states, $U \subseteq \mathbb{R}^m$ is the set of possible control inputs, $P$ is set of modes that the car may be in, $f: \mathbb{R}^n \times \mathbb{R}^m \rightarrow \mathbb{R}^n$ is a Lipschitz continuous function that we call the {\em dynamics} function, and $g: X \times P \rightarrow U$ is the {\em controller} function.

%Given an initial state $x_0$ A trajectory of the system is a function $\xi: X \times U \times P \times \mathbb{R}^+ \rightarrow X$, where
%$$ \frac{d\xi(x,g(x),t)}{dt} = f(\xi(x, g(x), t), t).$$

% %how is position, model, uncertainty disturbances related to risk?
%how are the control inputs handeled?
% \sayan{New text}
In this section, we discuss how our predictive online reachability analysis module works to inform the  controller, and helps avoid collisions with pedestrians.

\subsection{Online Reachability}
%\scriptsize

Recall from Section~\ref{sec:background}, that for online safety verification or monitoring, the  estimated current states $\Theta$ of the vehicle and the pedestrian(s) are propagated forward in time for a look-ahead period $T_{\mathit{Look}}$, according to the model $\A$, to check whether there can be a loss of safe separation within that time.
%
%
%This is an instance of the  {\em reachability analysis problem\/} which is the cornerstone of formal verification~\cite{alur95algorithmic,girard2005reachability}. Thus far, most applications of reachability analysis for robotic and cyber-physical systems have been for offline design verification. 
For this to be effective, the online reachability analysis must be: (1) {\em accurate}---to reduce false positives, and (2) {\em quick}---to allow for timely recovery and mitigating actions.
% \begin{itemize}
%     \item {\em accurate}---to reduce false positives, 
%     \item {\em quick}---to allow for timely recovery and mitigating actions, if needed.
% \end{itemize}
Specifically, for $T_{\mathit{Look}} = 3$ to $5$ seconds look-ahead,
the reachability analysis must finish within $10$-$100$ milliseconds.
Current offline reachability algorithms for hybrid systems need minutes of computation time for 6 to 12-dimensional nonlinear models.
Making the problem more challenging, the uncertainties in the  initial position, heading, and speed, defining the set $\Theta$ may be considerable because of estimation errors (see Sec. \ref{sec:vehicle-hardware}). The larger these errors are, the larger $\Theta$ needs to be to achieve the same level of confidence in that it contains the actual initial state of the car, and consequently achieve the same level of confidence that the corresponding reach set contains the future state of the car. 
%\sayan{(lets give a number/range)}, 
Further, there may be multiple pedestrian and vehicle trajectories to propagate forward and multiple modes to evaluate. 
 % , such as $\brake$ and $\track$
 %
%  Thus, for scalable online monitoring, a 3 orders of magnitude improvement in verification time is needed. 
 In this paper, we identify  a restricted class of scenarios, and for those, we show how this is  achieved in our \reachmodule.
 
%
% Impact
% (c) In the  context of automotive systems, {\em risk\/} is  broadly defined as: {\em severity of accident $\times$ probability of its occurrence}~\cite{simonepresent}. We have shown earlier how reachability analysis can be used for off-line risk analysis~\cite{FanQM:IEEEDT2018}; online risk analysis will not only be an attractive  new application for our new reachability technologies, it will also aid bandwidth allocation for BEV construction. 

%\todo{What are the modes? Brake, cruise, accelerate? Connect it with the actual model.}
%Section on DryVR and reachability pipeline
%\todo{Do we need to talk about hybrid here? Otherwise just present as if $\A$ is a dynamical system and later say that the approach can also handle hybrid and give references.}
Given a dynamical system $\A$ over a state space $\X$ and having an input space $\U$, a mode space $\P$, a dynamics function $f: \X \times \U \rightarrow \X$, a closed loop controller function $g: \X \times \P \rightarrow \U$, a set of possible initial states $\Theta \subseteq \X$, 
%representing the uncertainty in its state estimate, 
a mode $p \in P$, and a look-ahead time $T>0$, a state $\vx \in \X$ is reachable if there exists an execution of $\A$ starting from $\Theta$, having a fixed mode $p$, and of duration at most $T$, that reaches $\vx$. The set of reachable states is written as $\reach{\A}(\Theta,p, T) \subseteq \X$. 
%
% Sayan: choosing not to give formal definitions here. 
% we are out of space.
Our design of \reachmodule uses the data-driven reachability approach as implemented in the DryVR~\cite{FanQM18} and C2E2~\cite{DMVemsoft2013} tools.  Roughly speaking, simulation data generated from a model or an executable for $\A$, is used together with sensitivity analysis of $\A$ to compute over-approximations of $\reach{\A}(\Theta,p, T)$.
We are able to achieve the online-level computation times (as shown in Table~\ref{table:reachsettimes}) by using a few tricks that may be broadly applicable: 
(a) we fixed the function $\beta$ quantifying sensitivity off-line, 
(b) we capped the number of simulation traces used---this sacrifices precision for lowering computation, 
(c) we use low-dimensional dynamical vehicle models that are adequate for simple safety requirements like separation. The loss of accuracy from these approximations is partly mitigated by computing multiple reach sets from  different initial sets corresponding to different levels of uncertainty or confidence in state estimation. 
%levels.
%\todo{``confidence level'' took me by surprise... do we introduce this earlier somewhere?}
%learning the sensitivity function off-line using DryVR
%\todo{In the previous para we said that a orders of magnitude improvement is needed. Here, give some more intuition about why this approach works. Offline sensitivity analysis? How? Low dimensional reach set? }

In more detail, our reachability analysis procedure is adopted from the algorithm developed in~\cite{FanQMV:CAV2017}. 
%
%given the properties of the sensors and the uncertainty in the perception module, the state of the car is first estimated to be in a ball $\Theta$ of radius $r$ and center $\vx_0$ in $\X$. Moreover, a mode $p \in \P$ that we need to evaluate its safety, is given. 
%
First, in an off-line procedure, the sensitivity function $\beta$ of the system is learned from sampling system trajectories. This function bounds the distance between trajectories (or solutions) of $\A$ starting from different initial states. 
For this step, we use the DryVR tool of~\cite{FanQMV:CAV2017}.

For the online computation, consider an initial state estimate $\Theta$ with radius $r$. This set is partitioned into $m$ regions. For each region $i$, a numerical simulation $\xi(\vx_i,t)$ of $\A$ starting from a representative state $\vx_i$  is computed up to the look-ahead time $T_{Look}$.
The over-approximation of the reach set $\reach{\A}(\Theta,p,T_{\mathit{Look}})$ is computed by {\em bloating} each simulation $\xi(\vx_i,t)$ with $\beta$, and then taking the union of all these sets.
It has been shown in earlier works that~\cite{DMVemsoft2013,breachAD}, provided the sensitivity function is accurate, the computed reachable set can be made arbitrarily precise by increasing $m$.
The output $\reach{\A}(\Theta,p,T_{\mathit{Look}})$ is used to detect separation with the pedestrian's predicted paths by the decision module.
Different $r$ can be chosen based on the desired level of confidence. If we want a higher level of confidence, a larger $r$ will be preferable, which will increase the covered region of the initial set and consequently, the reach set $\reach{\A}(\Theta,p,T_{\mathit{Look}})$. Thus, the risk of actual trajectory of the vehicle not being covered by the reach set will be mitigated.

%in line~\ref{ln:simulate}
% After that, using a function $\beta: \mathbb{R}^+ \times \mathbb{R}^+ \rightarrow \mathbb{R}^+$ quantifying the sensitivity of the dynamics to initial states, along with the computed simulation, an
% % the algorithm computes the 
% over-approximation of the reach set is computed.
%and returns it in line~\ref{ln:bloat}. 
% Given an initial set of initial states $\Theta$ with radius $r$, and a time instant $t$, $\beta(r,t)$ is an upper bound on how far is the actual state of the car from the simulated state at time $t$.
%The function ``bloat'', returns a sequence of balls in $\X$. For the state $\xi(t)$ of the simulation $\xi$ at time $t$, ``bloat'' constructs a ball centered at $\xi(t)$ with radius $\beta(r, t)$.
% The over-approximation of the reach set is computed by {\em bloating} the computed simulation using $\beta$. For the state $\xi(t)$ of the simulation $\xi$ at time $t$, bloating means constructing a ball centered at $\xi(t)$ with radius $\beta(r, t)$. The tool DryVR would compute several simulations from randomly chosen states from the initial set $\Theta$ to estimate $\beta$, partition the initial set and simulate again if the resulting quality of the reach set needs improvement, and use the actual model of the car which can be very complex, high dimensional, and/or not available for the safety engineers. 

%\todo{Justify this model in a sentence.}
\paragraph*{Vehicle model}
We use the standard 5-dimensional bicycle model for the vehicle, which has been shown to accurately forecast an autonomous vehicle state \cite{kong2015kinematic},  where
%shown in the following paragraph for our safety checking. Moreover, we estimate its $\beta$ off-line instead of learning it every time a reach set needs to be computed.
%As discussed, we use a simplified 5-dimensional bicycle dynamical model 
$f$ is given by:
\begin{align*}
\label{eq:dubin}
\dot{x} = v \cos \theta, \; \dot{y} = v \sin \theta, \; \dot{\phi} = u, \; \dot{v} = a, \; \dot{\theta} = \frac{v}{L} \tan(\phi) 
\end{align*}
where $u$ is the input steering angular velocity and $a$ is the input acceleration that are determined by the velocity and waypoint following controller $g$ of Section~\ref{sec:control}, $(x, y)$ is the position, $v$ is the speed, $\theta$ is the  steering angle, $\phi$ is its heading angle and $L$ is the length of the car. Hence, the state space $\X = \mathbb{R}^5$, the control space $\U = \mathbb{R}^2$, and the set of modes is $\{\track, \brake\}$. The mode affects the dynamics indirectly by determining the control inputs computed in Section~\ref{sec:control}. In the $\brake$ mode, we set the reference speed to zero and apply the resulting PID control, as there is no brakes to apply in the bicycle model as opposed to the real car.

\paragraph*{State estimation uncertainties}
There are several sources of errors and uncertainties in measuring the initial state of the above system. For example, the position of the car and the pedestrian have uncertainties arising from the localization system. 
%All these are 
We define a nested sequence of  initial sets $\Theta_1 \subset \Theta_2 \subset \ldots \Theta_k$---increasingly conservative and with increasing levels of confidence. For each $\Theta_i$ we  compute corresponding over-approximations of $\reach{\A}(\Theta_i,p, T)$ using the above procedure starting from $\Theta_i$. From the monotonicity of $\reach{\A}$, it follows that 
$\reach{\A}(\Theta_1,p, T) \subseteq \reach{\A}(\Theta_2,p, T) \subseteq \ldots \reach{\A}(\Theta_k,p, T)$, and therefore, if $\reach{\A}(\Theta_i,p, T)$ is verified safe, then so are the smaller sets. Thus, if $\reach{\A}(\Theta_i,p, T)$ is verified safe in the \decisionmodule, then computation stops. 
%Further, When computing any of these sets, we set the number of refinements according to the partition budget. 
%
%\sayan{Revisit this after figures and metrics are finalized.}
%Thus, a single set is computed from the specified initial set. This is reasonable since the error in the sensors and the time interval over which we are verifying are small. 

\subsection{Evaluation of online reachability}
Implementing these strategies with a look-ahead time of $T_{\mathit{Look}}=3$ seconds gave us a maximum reachability computation time of only $0.1$ seconds on the vehicle's computers as shown in Table~\ref{table:reachsettimes}. We can see as well, from the same table, that the computation time of reach sets %Algorithm~\ref{fig:reachability} 
increases linearly with the look-ahead time. 
This shows the promise of our approach particularly because our current implementation of \reachmodule does not yet exploit parallelism of the simulation step of the algorithm. 

The results in Table~\ref{table:reachsetacc} show that the bicycle vehicle dynamics and the reach set obtained
%from Algorithm~\ref{fig:reachability} 
accurately predict the behavior of the real car. Even with the lowest confidence reach sets obtained from the smallest initial sets considered and longest look-ahead time, we acheived 92\% accuracy, while the maximum is about  99\%. To calculate the accuracy of the reach sets generated by the \reachmodule, we replay several simulations offline and check at each timestep $t$ of the simulation whether the reach set contains the real trajectory of the vehicle. We consider the computed reach set to be ``accurate'' for a given timestep $t$ if this is true.
%The rate at which we compute the reachset is also restricted by the time needed to get a new measurement state which is restricted in turn by the frame rate and the perception module computation time and the ROS communication time.  

\newcommand\tspace{\rule{0pt}{2.6ex}}         % = `top' strut
\newcommand\bspace{\rule[-0.9ex]{0pt}{0pt}}   % = `bottom' strut
\begin{table}[!t]
    \begin{center}
        \caption{Reach Set Computation Time}
        \label{table:reachsettimes}
        \begin{tabular}{@{}l|c|c|c|c|c@{}}
            \toprule
            T$_{Look}$ (s) & 3.0 & 3.5 & 4.0 & 4.5 & 5.0 \tspace\bspace\\
            % \midrule
            Compute Time (s) & 0.096 & 0.103 & 0.129 & 0.136 & 0.163 \tspace\bspace\\
            \bottomrule
        \end{tabular}
    \end{center}
    % \vspace{-0.5cm}
\end{table}
% \vspace{-0.5cm}
\begin{table}[!t]
    \begin{center}
        \caption{Reach Set Accuracy}
        \label{table:reachsetacc}
        \begin{tabular}{@{}l|c|c|c|c|c@{}}
            \toprule
            T$_{Look}$ (s) & 3.0 & 3.5 & 4.0 & 4.5 & 5.0 \tspace\bspace\\
            \hline
            High Conf. (\%) & 98.905 & 98.338 & 98.617 & 97.825 & 96.851 \tspace\bspace\\
            Med Conf. (\%) & 98.161 & 96.629 & 97.138 & 95.917 & 94.874 \tspace\bspace\\
            Low Conf. (\%) & 95.825 & 94.078 & 94.672 & 93.416 & 92.078 \tspace\bspace\\
            \bottomrule
        \end{tabular}
    \end{center}
\end{table}

\subsection{Decision Module}
\label{sec:decision}
For any initial set of states $\Theta$, mode $p$, and look-ahead time $T_{\mathit{Look}} > 0$, if the intersection $R \cap U$ of an over-approximation $R \supseteq \reach{\A}(\Theta,p,T_{\mathit{Look}})$ and an unsafe set $U$, is empty, then we can safely decide that the system is safe up to time $T_{\mathit{Look}}$. On the other hand, if $R \cap U \neq \emptyset$ then we cannot infer that there necessarily exists an unsafe execution.

The decision module ($\decisionmodule$) runs the reachability module \reachmodule 
%Algorithm~\ref{fig:reachability} 
on the mode $\track$ and checks if the resulting over-approximation of the reach set $R$ is safe when the unsafe set $U$ is the union of the reach sets of all surrounding pedestrians.
Their reach sets are computed by bloating
% applying Algorithm~\ref{fig:reachability} with
% the input simulation being
the trajectories obtained from the \piemodule module
% and the sensitivity function $\beta$ being 
using a constant sensitivity function equal to $r$, where $r$ is the radius of the initial set representing the uncertain position of the pedestrian.
If $R \cap U \neq \emptyset$, the decision module $\decisionmodule$ sets the mode for the controller to $\brake$ to avoid possible collision. Otherwise, it sets it to $\track$.

The decision module $\decisionmodule$ can run the reachability module \reachmodule
% Algorithm~\ref{fig:reachability} 
on mode $\brake$ as well, to check if there is a possibility of unavoidable collision, and alert the pedestrian. It can also compute the reach sets using initial sets with different radii to get different levels of confidence in safety.

Lastly, the results from Tables~\ref{table:reachsettimes} and~\ref{table:reachsetacc} show that \decisionmodule can operate accurately in real time to ensure safety while preserving a smooth ride with little unnecessary braking.

%\paragraph*{Predictive decision making}
%How does decision making use Reach? 
%Time for finishing execution of Algorithm 1. 
%Forward pointer to experimental evaluation Table~\ref{}.

%\paragraph*{Safety assertions}
%Proposition. Assuming that localization error is bounded by $r$ and that $\A$ is an conservative approximation of the real model, reach(T) contains actual position of the vehicle.
 
%%%%%%%%%%%%%%%%%%%%%%%%%%%%%%%%%%%%%%%%%%%%%%%%%%%%%%%%%%%%%%%%%%%%%%%%%%%%%%%%
\begin{figure*}[!ht]
\centering
\hspace{0.3cm}\includegraphics[width=0.98\textwidth]{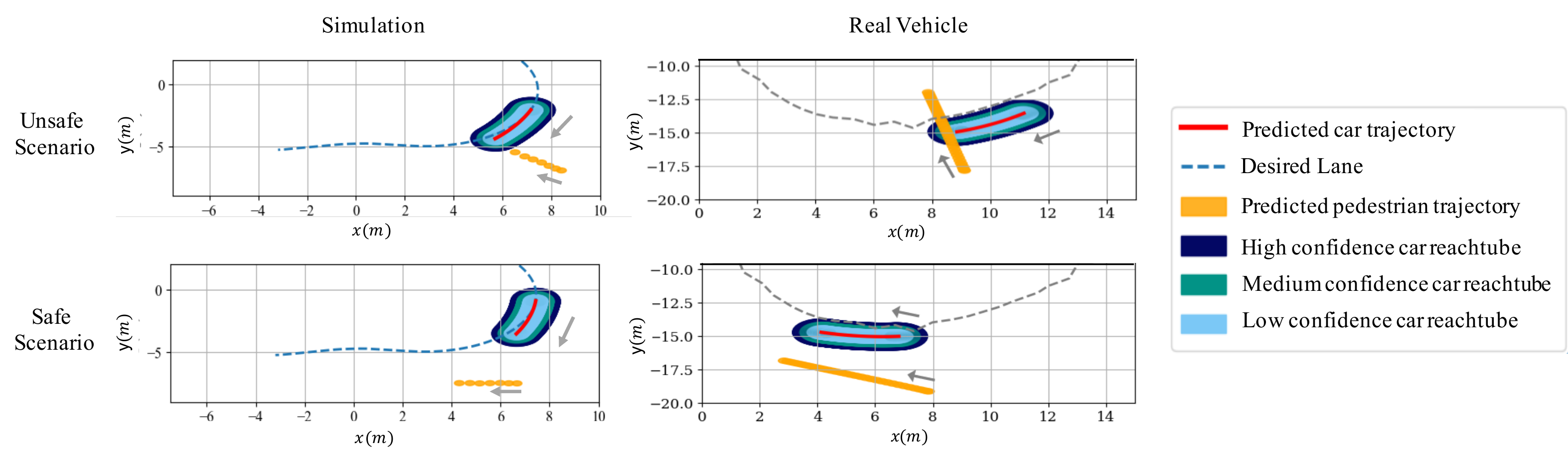}
\captionsetup{justification=centering}
\caption{Experimental results showing predicted pedestrian trajectories and vehicle reachtubes. Left two graphs show experiments done in simulation. Right two graphs show experiments performed on real vehicle. 
}
\label{fig:results} 
\vspace{-0.15cm}
\end{figure*}

\section{Experimental Results}
\label{sec:results}

In this section, we present the results of our monitoring system,  both in simulation and on our test vehicle. 

\subsection{Simulation}
\begin{figure}[!h]
	\centering
	\includegraphics[width=0.75\columnwidth]{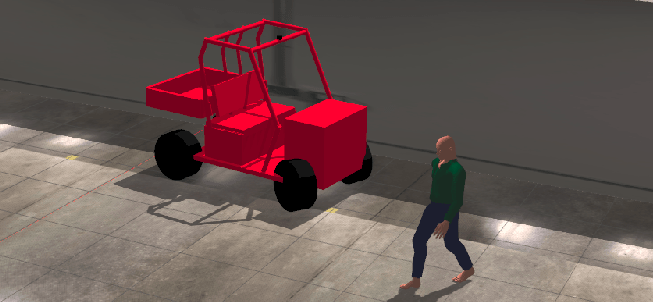}
	\caption{Gazebo simulation environment}
	\label{fig:gazebo_sim}
\end{figure}

We created a simulation environment in Gazebo (shown in Fig. \ref{fig:gazebo_sim}) to test our vehicle controllers and online reachability analysis module before deployment on a physical system.\footnote{Gazebo vehicle model adapted from \cite{polarismodel}.} The scenario used consists of a vehicle following a curved path with a pedestrian walking in close proximity.

Fig. \ref{fig:results} shows the results of the reachability analysis computed by the \reachmodule. It generates three sets of reach sets corresponding to high, medium, and low confidence levels. The different levels of confidence can be used to encode the uncertainty associated with the vehicle's starting state. The graphs in the upper and lower left hand side of Fig. \ref{fig:results} show simulation results. In the first case, the pedestrian moves parallel to the vehicle's path and the \reachmodule predicts that the future trajectory of the vehicle will be safe and allows it to continue. In the second case, the pedestrian crosses the vehicles path and the \reachmodule sends a signal to the controller telling the vehicle to brake.

\subsection{Real-World Testing}

% \begin{figure}[!h]
% 	\centering
% 	\includegraphics[width=0.90\columnwidth]{Figures/real_exp.pdf}
% 	\captionsetup{justification=centering}
% 	\caption{\textbf{Real vehicle experiment results}. Top: Safe scenario. Bottom: Unsafe scenario. Grey arrows indicate direction of motion.}
% 	\label{fig:real_reachsets}
% \end{figure}

To demonstrate our complete pipeline (with off the shelf components and in-house autonomous modules), we run our autonomous system in two test cases.
Fig. \ref{fig:results} shows two examples of the our integrated system on a real vehicle. In both cases, as the vehicle comes around the curved lane in the test arena, a pedestrian is detected and tracked.

The \piemodule module estimates the goal location and passes the predicted trajectory to the \reachmodule.
In the lower right portion of Fig. \ref{fig:results}, the predicted trajectory shows the pedestrian walking near the lane without intersecting the vehicle's path. As a result, the vehicle proceeds to safely move past the pedestrian without braking at any point during the experiment. 
In the upper right portion of Fig. \ref{fig:results}, the predicted trajectory suggests that the pedestrian will cross the lane. The \reachmodule decides this situation is unsafe as a collision is likely and signals the controller to brake.

The \reachmodule gets estimates of the pedestrian's intent and vehicle state and performs reachability analysis to infer whether the system is safe up to a look-ahead time $T_{\mathit{Look}}$. As discussed earlier, the state estimates can have large and fluctuating errors and thus we use nested uncertainty bounds with different levels of confidence. These confidence levels are illustrated in the high, medium, and low confidence reach sets in Figure \ref{fig:results}. Depending on the level of risk one can tolerate, the decision to brake can be made using a higher or lower confidence reach set. 

Similar to what we observed in simulation, our experiments on a real vehicle, for a look-ahead time of $T_{\mathit{Look}}=3$ seconds, saw reachability computation times of less than $0.1$ seconds on average. Increasing the number of reach sets by using initial sets with different confidence levels or allowing reach set refinements can give more precise monitoring at the cost of increased computation. 

\section{Discussion}
\label{sec:discussion}

We present an integrated autonomous system that uses a novel pedestrian intent estimator to safely maneuver amongst humans.
% We present a pedestrian intent estimation framework that can accurately predict future pedestrian trajectories given multiple possible goal locations.
We added another layer of safety and risk assessment by developing a reachability-based online monitoring scheme that formally assesses the safety of these interactions with nearly real-time performance ($\sim0.1s$).
These techniques are tested in simulation and integrated on a test vehicle with a complete in-house autonomous stack, demonstrating effective and safe interaction in real-world experiments. 

In our current experiments, we make many assumptions and control many aspects of the system to ensure the baseline performance is functional.  
However, in reality, any of the submodules may fail.
For example, we only tested scenarios where the \piemodule correctly predicted the pedestrian.  
Given that no model can perfectly predict a human, we hope to explore scenarios where the prediction is incorrect, making the need for an online monitor more pressing.
We note that similar statements hold for the vision and localization modules.
We also found that timing and synchronization of the components were difficult and had a noticeable impact on the safety assessment.
We hope to include such inaccuracies and uncertainties in future iterations of our online monitor.

%%%%%%%%%%%%%%%%%%%%%%%%%%%%%%%%%%%%%%%%%%%%%%%%%%%%%%%%%%%%%%%%%%%%%%%%%%%%%%%%

% ref: \cite{IEEEexample:articleetal}, \cite{IEEEexample:book}, ...

\bibliographystyle{IEEEtran}
\bibliography{IEEEabrv,IEEEexample,kt-bib,sayan1}

\end{document}